\documentclass{article}

\usepackage{microtype}
\usepackage{graphicx}
\usepackage{subfigure}
\usepackage{booktabs} 

\usepackage{hyperref}

\usepackage[accepted]{icml2025}

\usepackage{amsmath}
\usepackage{amssymb}
\usepackage{mathtools}
\usepackage{amsthm}

\usepackage[capitalize,noabbrev]{cleveref}

\theoremstyle{plain}
\newtheorem{theorem}{Theorem}[section]

\theoremstyle{definition}

\theoremstyle{remark}

\newtheorem{example}[theorem]{Example}

\usepackage[textsize=tiny]{todonotes}

\usepackage{physics}  
\usepackage{commath}  
\usepackage{multirow}  

\newcommand{\bx}{\mathbf{x}}
\newcommand{\bW}{\mathbf{W}}
\newcommand{\ie}{\textit{i}.\textit{e}.}

\def\secref#1{Section~\ref{#1}}
\def\tabref#1{Table~\ref{#1}}
\def\eqref#1{Eq.~\ref{#1}}

\DeclareMathOperator{\logit}{logit}
\DeclareMathOperator\arctanh{arctanh}
\DeclareMathOperator{\softmax}{softmax}

\usepackage[most]{tcolorbox}

\tcbset {
  base/.style={
    arc=0mm,
    bottomtitle=0.5mm,
    boxrule=0mm,
    colbacktitle=black!10!white,
    coltitle=black,
    fonttitle=\bfseries,
    left=2.5mm,
    leftrule=1mm,
    right=3.5mm,
    title={#1},
    toptitle=0.75mm,
  }
}

\definecolor{brandblue}{rgb}{0.34, 0.7, 1}
\newtcolorbox{mainbox}[1]{
  colframe=brandblue,
  base={#1}
}

\newtcolorbox{subbox}[1]{
  colframe=black!30!white,
  base={#1}
}

\definecolor{codegreen}{rgb}{0,0.6,0}
\definecolor{codegray}{rgb}{0.5,0.5,0.5}
\definecolor{codepurple}{rgb}{0.58,0,0.82}
\definecolor{backcolour}{rgb}{0.95,0.95,0.92}

\lstdefinestyle{mystyle}{
    backgroundcolor=\color{backcolour},
    commentstyle=\color{codegreen},
    keywordstyle=\color{magenta},
    numberstyle=\tiny\color{codegray},
    stringstyle=\color{codepurple},
    basicstyle=\ttfamily\footnotesize,
    breakatwhitespace=false,
    breaklines=true,
    captionpos=b,
    keepspaces=true,
    numbers=left,
    numbersep=5pt,
    showspaces=false,
    showstringspaces=false,
    showtabs=false,
    tabsize=2
}

\icmltitlerunning{Temperature-Free Loss Function for Contrastive Learning}

\begin{document}

\twocolumn[
	\icmltitle{Temperature-Free Loss Function for Contrastive Learning}



	\icmlsetsymbol{equal}{*}


	\begin{icmlauthorlist}
		\icmlauthor{Bum Jun Kim}{yyy}
		\icmlauthor{Sang Woo Kim}{yyy}
	\end{icmlauthorlist}

	\icmlaffiliation{yyy}{Department of Electrical Engineering, Pohang University of Science and Technology, Pohang, South Korea}

	\icmlcorrespondingauthor{Sang Woo Kim}{swkim@postech.edu}

    \icmlkeywords{Machine Learning, ICML}

	\vskip 0.3in
]



\printAffiliationsAndNotice{}  

\begin{abstract}
	As one of the most promising methods in self-supervised learning, contrastive learning has achieved a series of breakthroughs across numerous fields. A predominant approach to implementing contrastive learning is applying InfoNCE loss: By capturing the similarities between pairs, InfoNCE loss enables learning the representation of data. Albeit its success, adopting InfoNCE loss requires tuning a temperature, which is a core hyperparameter for calibrating similarity scores. Despite its significance and sensitivity to performance being emphasized by several studies, searching for a valid temperature requires extensive trial-and-error-based experiments, which increases the difficulty of adopting InfoNCE loss. To address this difficulty, we propose a novel method to deploy InfoNCE loss without temperature. Specifically, we replace temperature scaling with the inverse hyperbolic tangent function, resulting in a modified InfoNCE loss. In addition to hyperparameter-free deployment, we observed that the proposed method even yielded a performance gain in contrastive learning. Our detailed theoretical analysis discovers that the current practice of temperature scaling in InfoNCE loss causes serious problems in gradient descent, whereas our method provides desirable gradient properties. The proposed method was validated on five benchmarks on contrastive learning, yielding satisfactory results without temperature tuning.
\end{abstract}

\section{Introduction}
Deep learning has shown remarkable results across numerous tasks using large datasets, often with labels. To cope with extensive manual labeling, research in self-supervised learning has sought to learn a representation of data without their labels \citep{DBLP:conf/nips/CaronMMGBJ20,DBLP:conf/nips/GrillSATRBDPGAP20}. Among several directions in self-supervised learning, recent studies on contrastive learning have demonstrated promising results in various fields, including computer vision, graph representation learning, and anomaly detection \citep{DBLP:conf/icml/ChenK0H20,DBLP:journals/corr/abs-2006-04131,DBLP:conf/aaai/ReissH23}.

\begin{table}[t]
	\caption{Numerous studies with InfoNCE loss have adopted their own temperature values}
	\label{tab:temp}
	\begin{center}
		\begin{small}
			\begin{sc}
				\begin{tabular}{l|c}
					\toprule
					Study                                    & Best Temperature \\
					\midrule
					\citet{DBLP:conf/cvpr/He0WXG20}          & 0.07             \\
					\citet{DBLP:conf/cvpr/WuXYL18}           & 0.07             \\
					\citet{DBLP:conf/nips/CaronMMGBJ20}      & 0.1              \\
					\citet{DBLP:conf/aaai/ReissH23}          & 0.25             \\
					\citet{DBLP:conf/nips/GrillSATRBDPGAP20} & 0.3              \\
					\citet{DBLP:conf/icml/ChenK0H20}         & 0.1 or 0.5       \\
					\bottomrule
				\end{tabular}
			\end{sc}
		\end{small}
	\end{center}
	\vskip -0.1in
\end{table}

Contrastive learning aims to learn a representation space of data using the similarity scores of possible pairs that are obtained through different views \citep{DBLP:journals/pami/DosovitskiyFSRB16}. A common approach in contrastive learning is to adopt InfoNCE loss \citep{DBLP:journals/corr/abs-1807-03748,DBLP:conf/icml/ChenK0H20}, whose minimization enables discriminating similarities and dissimilarities within pairs, thereby acquiring representation of data.

\begin{figure*}[t!]
	\begin{center}
		\centerline{\includegraphics[width=\linewidth]{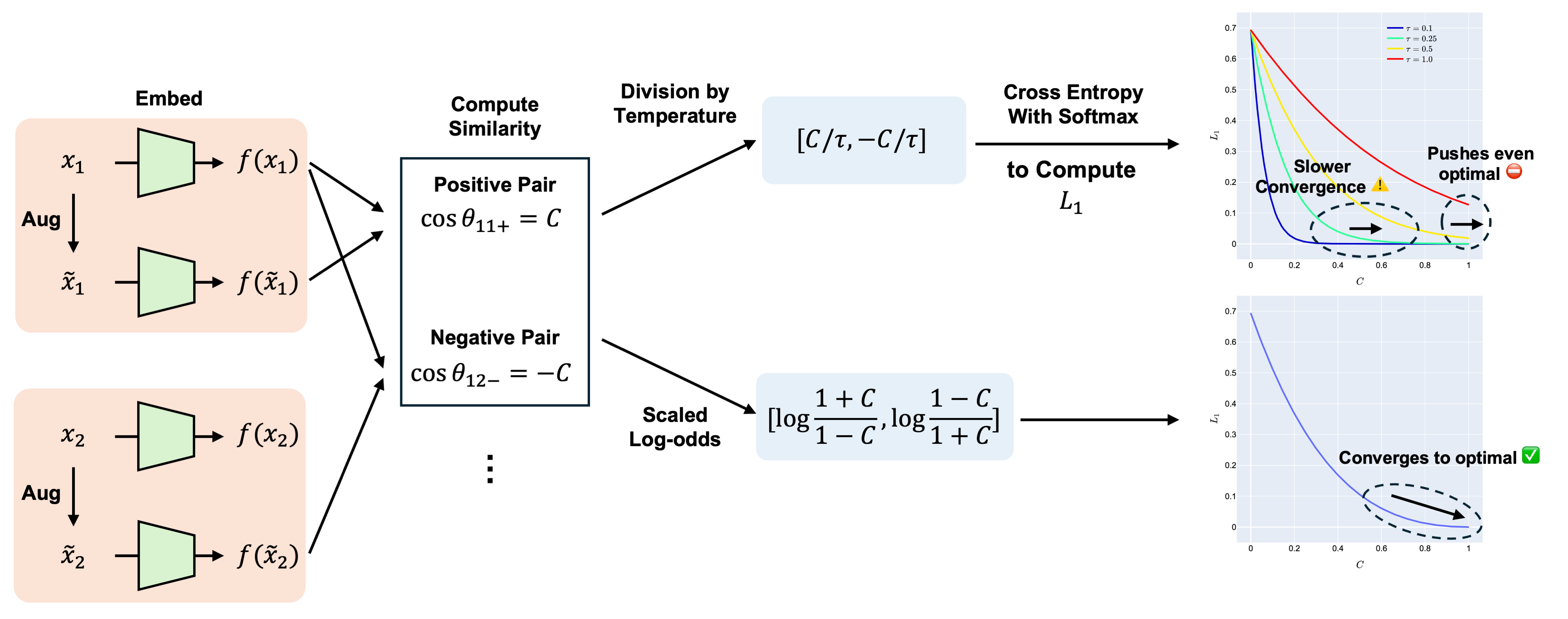}}
		\caption{Overview of contrastive learning with InfoNCE loss with existing division by temperature scheme and proposed method. The cosine similarities $[C, -C]$ follow the scenario in the main text. When approaching the optimal point $C \rightarrow 1$, the division by temperature scheme raises several problems in gradient descent, whereas the proposed method ensures convergence. The graphs on the right represent the loss $L_1$.}
		\label{fig:diagram}
	\end{center}
	\vskip -0.2in
\end{figure*}

The similarity score in InfoNCE loss is accompanied by a division by temperature to calibrate its scale. However, several studies have emphasized the significance of temperature to the performance of contrastive learning \citep{DBLP:conf/cvpr/WangL21a,DBLP:conf/iclr/ZhangZZPYK22}. Although the current consensus is to adopt a low temperature such as 0.07, 0.1, or 0.25, the exact value differs by each study (\tabref{tab:temp}). Indeed, numerous studies using InfoNCE loss have searched for the best temperature that is obtained through their own hyperparameter tuning via trial-and-error-based approaches. For example, MoCo \citep{DBLP:conf/cvpr/He0WXG20} used a temperature of 0.07, whereas MSC \citep{DBLP:conf/aaai/ReissH23} adopted 0.25. \citet{DBLP:conf/nips/GrillSATRBDPGAP20} reported that the temperature of 0.1 from \citet{DBLP:conf/icml/ChenK0H20} did not work suitably with their method and instead used 0.3 that is obtained from their hyperparameter search. However, in accordance with pre-existing hyperparameters, such as the learning rate and weight decay, finding the best combination of hyperparameters requires an exponentially large search space and experiments \citep{DBLP:conf/gecco/PfistererRPMB21,DBLP:conf/icml/FalknerKH18}. The difficulty in finding the best combination of hyperparameters becomes a further serious problem when training a large model \citep{DBLP:conf/nips/YangHBSLFRPCG21}.

To address the difficulty of temperature tuning, this study proposes a novel method to adopt InfoNCE loss without any temperature. Specifically, we replace the division by temperature with a scaled log-odds mapping, or equivalently the inverse hyperbolic tangent function. By injecting few lines of code into the existing design pattern of InfoNCE loss, we can enjoy temperature-free contrastive learning without extensive experiments to search for the best temperature value. To the best of our knowledge, this study is the first in a direction to completely remove the division by temperature scheme in InfoNCE loss to eliminate the need to search for its valid value.

Apart from the temperature-free property, we observed that the proposed method also brought performance gains in contrastive learning. Through theoretical analysis, we discover that the existing division by temperature scheme is prone to vanishing gradient, which hinders optimization. We detail this behavior and analyze why the minimization of the InfoNCE loss is significantly sensitive to temperature value (\secref{sec:prob}). By contrast, we prove that the proposed method ensures alive gradients during training, which facilitates gradient descent optimization (\secref{sec:met}). Our comprehensive experiments demonstrate that the proposed method---without temperature tuning---yields a promising performance that is comparable to, or even better than, the existing scheme of division by temperature (\secref{sec:exp}).

\section{Contrastive Learning With InfoNCE Loss}
\label{sec:cl}

\subsection{Background}
Given a mini-batch of data $\bx_i$ with $i \in \{1, 2, \cdots, N\}$, contrastive learning \citep{DBLP:journals/pami/DosovitskiyFSRB16} obtains their different views $\tilde{\bx}_i$ via transformation such as data augmentation (\figref{fig:diagram}). Using a model $f$ such as a deep neural network, their embedding vectors $f(\bx_i)$ and $f(\tilde{\bx}_i)$ are obtained.

The InfoNCE loss computes cross-entropy with softmax using the scaled similarity scores of possible pairs within a mini-batch \citep{DBLP:journals/corr/abs-1807-03748,DBLP:conf/icml/ChenK0H20}. The positive pair should exhibit higher similarity because the representations are obtained from merely different views of the same instance. By contrast, negative pairs should exhibit no similarity because they are arbitrary pairs within a mini-batch \citep{DBLP:conf/kdd/ChenSSH17}. In other words, the InfoNCE loss indicates cross-entropy $-\sum_i y_i \log{\hat{y}_i}$ that applies one-hot label $y_i$ describing the correctness of pairs and penalizes a case if $\hat{y}_i$, the softmax output from the scaled similarity score, is low for a positive pair. Learning this matching task enables the model to capture the nature of the dataset and preserved structure in data augmentation, thereby acquiring a generic representation of data \citep{DBLP:conf/eccv/JoulinMJV16}. Indeed, several studies have reported that the acquired representation from contrastive learning achieves outstanding performance, even when compared with counterparts from supervised learning; the learned representation has been reported to work with a simple classifier, such as linear or nearest neighborhood \citep{DBLP:conf/icml/ChenK0H20,DBLP:conf/cvpr/He0WXG20}.

The original InfoNCE loss, proposed by \citet{DBLP:journals/corr/abs-1807-03748}, compares the dot product of two vectors across a mini-batch. Additionally, \citet{DBLP:conf/cvpr/WuXYL18} incorporated a division by temperature after computing the dot product. Moreover, the representations $f(\bx_i)$ and $f(\tilde{\bx}_i)$ are usually constrained into a unit sphere through explicit $l_2$ normalization \citep{DBLP:conf/nips/Sohn16}. This restriction leads to computing the dot product from normalized vectors, which is equivalent to the cosine of the angle between the two vectors. Based on this observation, \citet{DBLP:conf/icml/ChenK0H20} formulated the normalized temperature-scaled cross-entropy loss, which they refer to as the NT-Xent loss. This loss is defined as $L \coloneqq \sum_i L_i$ for
\begin{align}
	L_i \coloneqq -\log{
		\frac{
			\exp(\cos{\theta_{ii+}}/\tau)
		}
		{
			\sum_{j} \exp(\cos{\theta_{ij}}/\tau)
		}
	},
\end{align}
where $\cos{\theta_{ij}} \coloneqq f(\bx_i)^\top f(\tilde{\bx}_j) / \norm{f(\bx_i)} \norm{f(\tilde{\bx}_j)}$ with angle $\theta_{ij}$ between the two vectors, $\tau$ is a temperature, and the index $ii+$ indicates a positive pair. We later denote the index $ij-$ as negative pairs of $j \neq i$. Although the exact definition of NT-Xent loss differs from that of the original InfoNCE loss by \citet{DBLP:journals/corr/abs-1807-03748}, modern literature on contrastive learning \citep{DBLP:conf/cvpr/He0WXG20} usually adopts NT-Xent loss as InfoNCE loss. Following this definition, in the remainder of this paper, we refer to NT-Xent loss as InfoNCE loss.

\subsection{Related Works}
One difficulty in adopting InfoNCE loss is to search for a valid temperature value $\tau$, which has been found to be crucial to the performance of contrastive learning \citep{DBLP:conf/cvpr/WangL21a,DBLP:conf/iclr/ZhangZZPYK22}. Owing to the complex dynamics of temperature in contrastive learning, \citet{DBLP:conf/cvpr/ZhangZPNQYK22} proposed using dual temperatures, which yet comes with extended difficulty in hyperparameter tuning. Rather than using a fixed value of temperature, \citet{DBLP:conf/iclr/KuklevaBSK023} scheduled the temperature with respect to epochs during training while predefining its maximum and minimum. \citet{DBLP:journals/corr/abs-2308-01140} and \citet{DBLP:conf/icml/HuangCW0L0C23} proposed an adaptive temperature strategy depending on the input but using other hyperparameters to determine its scale. Although these studies have addressed the complex dynamics of temperature in InfoNCE loss, they still require hyperparameter tuning, often with the increased number of hyperparameters.

Another notable direction is to set the temperature as a learnable parameter in gradient descent. Indeed, CLIP \citep{DBLP:conf/icml/RadfordKHRGASAM21} adopted a learnable temperature using an exponential function. \citet{DBLP:journals/corr/abs-2110-04403} cast temperature as uncertainty and set it to be learned as a function of the input. \citet{DBLP:conf/icml/QiuHYZ0Y23} claimed to apply individualized temperatures that are set to be learnable. Although the learnable temperature might converge to a proper value, we later discover that this approach brings another problem in practical gradient descent optimization (\secref{sec:prob}).

All of these studies have emphasized the difficulty of manual temperature search. Nevertheless, to the best of our knowledge, no study has challenged the removal of the division by temperature scheme itself to eliminate the need to search for a valid value of temperature, which is the main objective of this study. To this end, this study proposes a method to replace the division by temperature in the InfoNCE loss.

\section{Problem Statement}
\label{sec:prob}
Although the problem at hand is the difficulty in searching for a valid temperature, we claim that the division by temperature scheme also brings serious problems in gradient descent optimization.

\paragraph{Representational Limit in InfoNCE Loss}
Adopting the InfoNCE loss indicates penalizing the cross-entropy computed along with the softmax output using a one-hot label that assigns one to a positive pair and zeros to negative pairs. Specifically, the InfoNCE loss regards the logits---which are input to softmax---as temperature-scaled cosine similarities, $[\cos{\theta_{ij}}/\tau]_j$. Ideally, minimizing the InfoNCE loss leads to $\cos{\theta_{ii+}}=1$ for the positive pair and $\cos{\theta_{ij-}}=-1$ for all other negative pairs of $j \neq i$.

However, in principle, the input to the softmax has to be arbitrary real numbers without constraints on the value. For example, a classification task typically obtains logits from a fully connected layer before the softmax \citep{DBLP:conf/cvpr/HeZRS16}, whose values are on the interval $(-\infty, \infty)$ with infinite length. A constraint on values arises on the subsequent softmax to be on the interval $[0, 1]$. Nevertheless, the cosine itself is constrained to the interval $[-1, 1]$. The division by temperature maps cosine similarities into an interval $[-1/\tau, 1/\tau]$ with plausible length: For example, choosing a low temperature such as $\tau=0.1$ yields an interval $[-10, 10]$. Although division by temperature might enlarge the interval of logits to a sufficient length, we claim that the interval length is still bounded by a finite length of $2/\tau$.

Note that the softmax is shift-invariant as $\softmax(\bx+k)=\softmax(\bx)$ for a constant $k$ and logits $\bx$ \citep{DBLP:conf/icml/BlancR18,DBLP:conf/nips/LahaCAKSR18}. Hence, this study examines the length of the interval for logits, not the exact interval. The finite length of the interval for input to the softmax causes a representational limit, especially on the softmax output. We begin with a motivational example.

\begin{example}
	Imagine two cosine similarities $[\cos{\theta_{ii+}}, \cos{\theta_{ij-}}] = [1, -1]$ with one positive and one negative, which is on the optimal point for minimizing InfoNCE loss. In this scenario, the element corresponding to the positive pair in the softmax output is $\sigma(2\tau)$, where $\sigma(x)=1/(1+\exp(-x))$, the sigmoid function. This element has to ideally be one because we assume the scenario of the optimal point. However, the sigmoid outputs $\sigma(2)=0.88$ for $\tau=1$, which slightly deviates from one. To solve this representational limit, a sufficiently high temperature should be chosen. Although choosing a lower temperature of $\tau=0.1$ yields $\sigma(20)=0.99$, the deviation from one is unavoidable for the division by temperature scheme. In the next, we detail this behavior and discover another side effect of the division by temperature scheme.
\end{example}

\paragraph{Vanishing Gradient of InfoNCE Loss}
We claim that the division by temperature scheme is prone to vanishing gradient, which hinders optimization in practice.

\begin{example}
	Imagine two cosine similarities $[\cos{\theta_{ii+}}, \cos{\theta_{ij-}}] = [C, -C]$ with one positive and one negative, where $C \in (0, 1)$.\footnote{Owing to the shift-invariance of softmax, rather than allowing the two degrees of freedom, we cast the problem to a single degree of freedom with $C$. In softmax, only the distance from the similarity score of the positive pair matters, not the exact score. Here, $C$ indicates half the distance $C = (\cos{\theta_{ii+}} - \cos{\theta_{ij-}})/2$. Note that the input to softmax in this example is equivalent to the input to the sigmoid function. A behavior with multiple cosine similarities is provided later.} Minimizing the InfoNCE loss should approach the optimal point as $C \rightarrow 1$. In this scenario, the $i$th InfoNCE loss becomes
	\begin{align}
		L_i & = -\log{\frac{\exp(C/\tau)}{\exp(C/\tau) + \exp(-C/\tau)}} \\
		    & = \log({1 + \exp(-2C/\tau)}).
	\end{align}
	Gradient descent updates the weight $\bW$ of the embedding model using the gradient $\pdv{L}{\bW} = \sum_i \pdv{L_i}{\bW}$. By the chain rule, we have $\pdv{L_i}{\bW} =\pdv{L_i}{C} \pdv{C}{\bW}$, whose scale is regulated by a scaling factor of
	\begin{align}
		\abs{\pdv{L_i}{C}} = \frac{2/\tau}{1+\exp(2C/\tau)}, \label{eq:grad}
	\end{align}
	which is determined by $C$ and $\tau$.

	\begin{figure}[t!]
		\begin{center}
			\centerline{\includegraphics[width=\columnwidth]{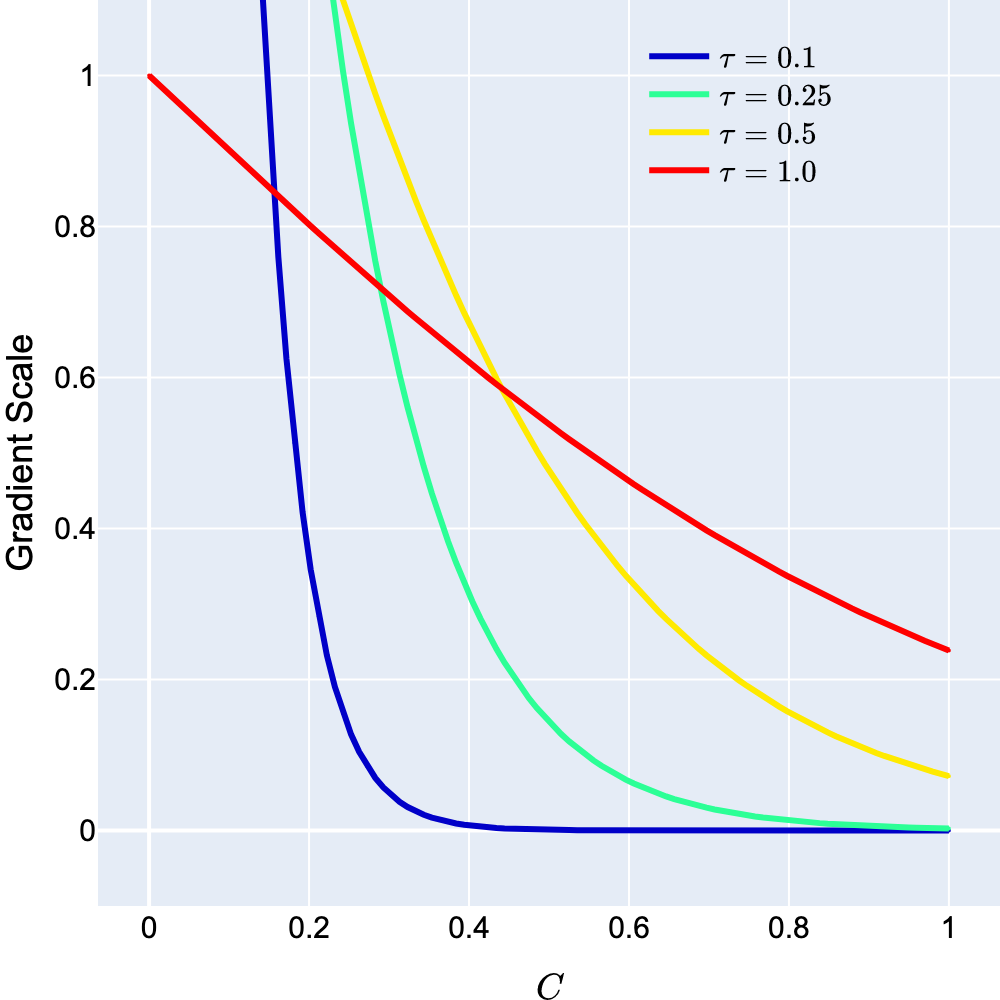}}
			\caption{Gradient scale corresponding to \eqref{eq:grad}. Higher temperatures yielded a nonzero gradient scale near the optimal point $C \rightarrow 1$. However, lower temperatures yield vanishing gradients at nonoptimal points.}
			\label{fig:grad}
		\end{center}
		\vskip -0.2in
	\end{figure}

	\begin{figure}[t!]
		\begin{center}
			\centerline{\includegraphics[width=\columnwidth]{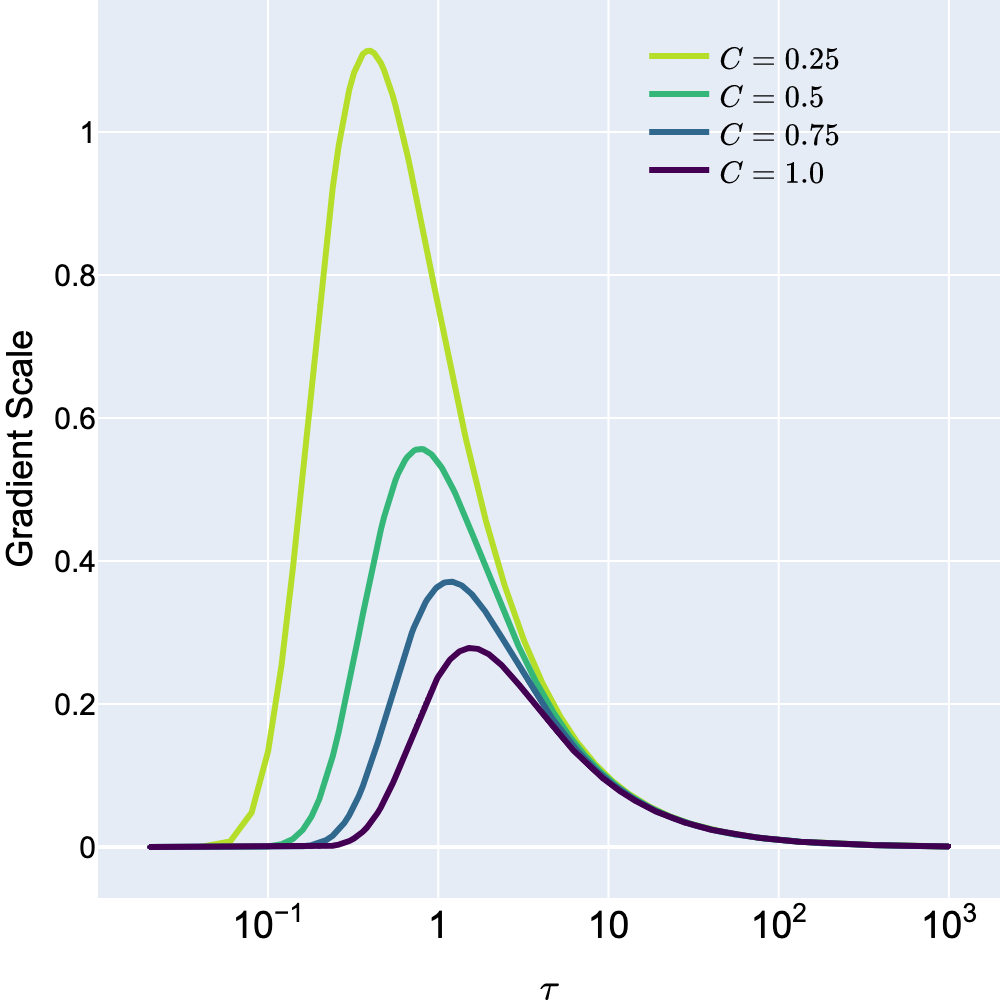}}
			\caption{Gradient scale corresponding to \eqref{eq:grad}. Lower $C$ should exhibit a nonzero gradient scale, but $C \rightarrow 1$ should yield a zero gradient scale; these conditions can only be satisfied for a precisely chosen temperature, such as $\tau=0.25$.}
			\label{fig:grad_tau}
		\end{center}
		\vskip -0.2in
	\end{figure}

	We plot this gradient scaling factor in \figref{fig:grad} for $\tau \in \{0.1, 0.25, 0.5, 1\}$. The optimal point $[\cos{\theta_{ii+}}, \cos{\theta_{ij-}}] = [1, -1]$ corresponds to $C=1$. However, at this optimal point, the gradient scale exhibits a nonzero value for high temperatures such as $\tau=1$ and $\tau=0.5$. The nonzero gradient at the minimum contradicts optimality and hinders convergence because it pushes the optimization further. This nonzero gradient arises from the representational limit of logits in the division by temperature scheme. Achieving a near-zero gradient at the optimal point requires a sufficiently low temperature, such as $\tau=0.1$ or $\tau=0.25$.

	However, a lower temperature brings a side effect of vanishing gradients. For $\tau=0.1$, the gradient scale becomes near-zero even on nonoptimal points in $C \in (0.4, 0.7)$, which we refer to as the vanishing gradient of the InfoNCE loss. This phenomenon slows convergence for minimizing the InfoNCE loss. Indeed, $C \in (0.4, 0.7)$ occurs in practice: \citet{DBLP:journals/corr/abs-2308-01140} reported that the distance of cosine similarity ranges from 0 to 1.4 in practice.

	\figref{fig:grad_tau} illustrates the gradient scale for $C \in \{0.25, 0.5, 0.75, 1\}$. Here, a smaller $C$ should exhibit a higher gradient scale, whereas $C \rightarrow 1$ should yield a near-zero gradient scale $\abs{\pdv{L_i}{C}} \rightarrow 0$. This criterion is satisfied only for a precisely chosen temperature value, such as $\tau=0.25$: Other choices degrade the optimization.
\end{example}

\paragraph{Further Difficulty of Tuning Temperature in Practice} We now consider multiple cosine similarities.

\begin{example}
	Imagine $N$ cosine similarities $[C, -C, \cdots, -C]$ with one positive and $N-1$ negatives, where $C \in (0, 1)$. Minimizing the InfoNCE loss should ideally approach the optimal point $C \rightarrow 1$. In this scenario, the $i$th InfoNCE loss becomes
	\begin{align}
		L_i & = -\log{\frac{\exp(C/\tau)}{\exp(C/\tau) + (N-1)\exp(-C/\tau)}} \\
		    & = \log({1 + (N-1) \exp(-2C/\tau)}),
	\end{align}
	with its gradient scale
	\begin{align}
		\abs{\pdv{L_i}{C}} & = \frac{(N-1)2/\tau}{(N-1)+\exp(2C/\tau)}, \label{eq:multi}
	\end{align}
	which is determined by $C$, $\tau$, and $N$.

	\begin{figure}[t!]
		\begin{center}
			\centerline{\includegraphics[width=\columnwidth]{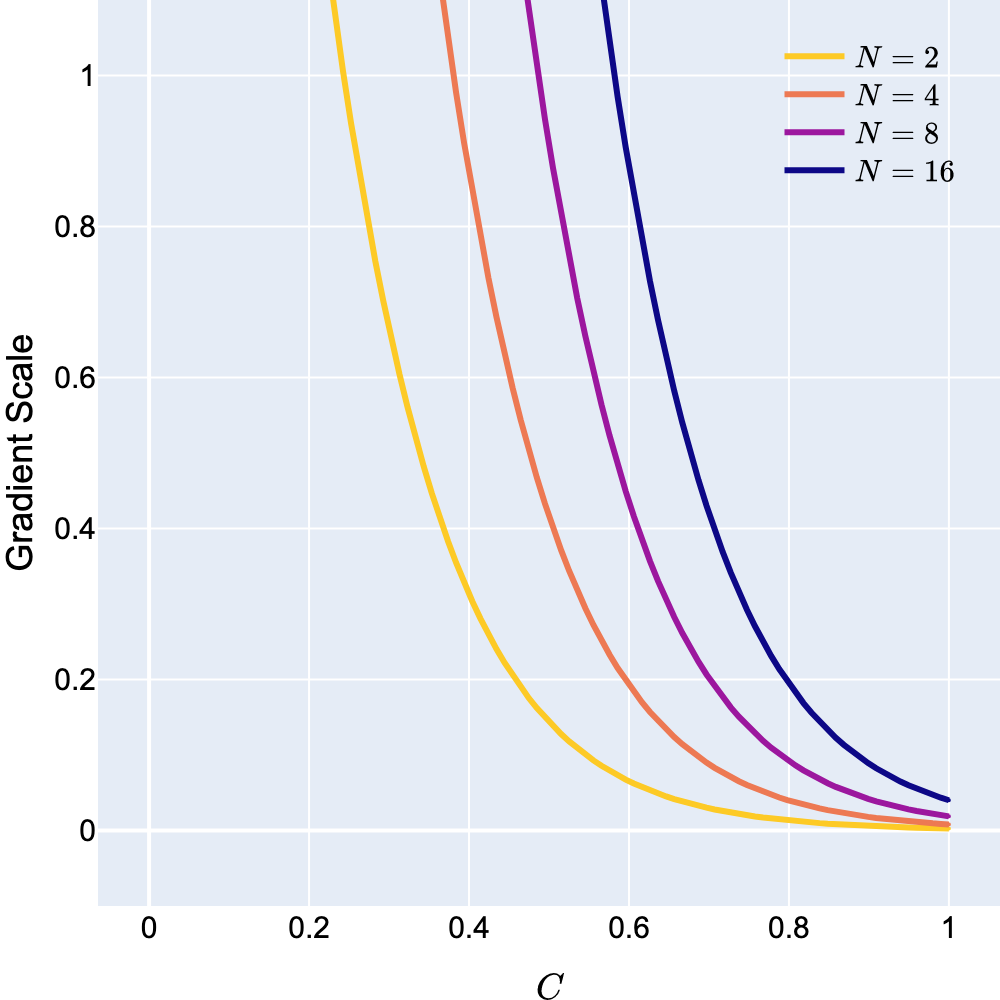}}
			\caption{Gradient scale corresponding to \eqref{eq:multi} for $\tau=0.25$. Although $N=2$ provides a near-zero gradient when approaching the optimal point $C \rightarrow 1$, other conditions such as $N=16$ yield a nonzero gradient scale, thereby affecting the valid temperature.}
			\label{fig:multi}
		\end{center}
		\vskip -0.2in
	\end{figure}

	\figref{fig:multi} illustrates the gradient scale for $C \in (0, 1)$, $\tau=0.25$ and $N \in \{2, 4, 8, 16\}$. Although we previously observed that $\tau=0.25$ provided an acceptable gradient property for $N=2$, this temperature does not work with other conditions, such as $N=16$, which exhibits a nonzero gradient when approaching the optimal point $C \rightarrow 1$. Again, we have to search for a valid temperature with desirable gradient properties for different conditions of $N$. This dependency on $N$ is one reason why the valid temperature differs with the target task: A universally applicable temperature does not exist. Later in \secref{sec:exp}, we experimentally demonstrate that the best temperature in practice differs for each target task.
\end{example}

\begin{subbox}{Summary of the Problem Statement}
	For the current scheme of division by temperature, we find a trade-off between 1) ensuring a near-zero gradient around the optimal point and 2) obtaining nonzero gradients on nonoptimal points. To achieve both properties, one has to precisely tune the temperature into a valid value. The gradient scale is even affected by other factors such as the number of pairs, which implies that a universally applicable value for temperature does not exist.
\end{subbox}

\paragraph{Vanishing Gradient of Learnable Temperature} \citet{DBLP:conf/icml/RadfordKHRGASAM21} set the temperature as a learnable parameter with gradient descent, which does not require hyperparameter tuning. Rather than division by temperature, they adopted multiplication of $\exp(t)$ on similarity scores with learnable parameter $t$. Although this approach might work in practice, we claim that it causes another gradient problem.

\begin{example}
	Imagine two cosine similarities $[\cos{\theta_{ii+}}, \cos{\theta_{ij-}}] = [C, -C]$ with one positive and one negative, where $C \in (0, 1)$. With the multiplication of $\exp(t)$, the $i$th InfoNCE loss becomes
	\begin{align}
		L_i & = -\log{\frac{\exp(C\exp(t))}{\exp(C\exp(t)) + \exp(-C\exp(t))}} \\
		    & = \log({1+\exp(-2C \exp(t))}).
	\end{align}
	Gradient descent updates the learnable parameter $t$ using the gradient $\pdv{L}{t} = \sum_i \pdv{L_i}{t}$, whose scale is regulated by
	\begin{align}
		\abs{\pdv{L_i}{t}} & = \frac{2C \exp(t)}{1+\exp(2C\exp(t))}.
	\end{align}
	Note that $C=0$ exhibits zero gradient scale for any $t$. This vanishing gradient is undesirable because it implies that any nonoptimal $t$ is subjected to no updates at $C=0$. Indeed, $C=0$ readily occurs in an initialized state, as observed in \citet{DBLP:journals/corr/abs-2308-01140}. Furthermore, while the aforementioned scenario corresponds to the multiplication of $\exp(t)$, the zero gradient at $C=0$ occurs for other variants such as the multiplication of $t$ or $1/t$.
\end{example}

\section{Proposed Method}
\label{sec:met}
We aim to build a loss function for contrastive learning that ensures alive gradients during training but zero gradient at the optimal point. Although staying at nonoptimal but moderate understanding of similarities such as $C=0.5$ might work in practice, we would like to enforce $C \rightarrow 1$ as much as possible to further capture similarities and dissimilarities in contrastive learning, thereby improving the quality of representation of data. The problem is closely related to the representational limit of logits, whose interval in the current scheme is bounded by the finite length of $2/\tau$. While preserving the existing structure of InfoNCE loss, we seek a valid mapping of cosine similarities.

Mapping a number from a finite interval to an infinite one has been addressed in logistic regression in the machine learning community for several decades \citep{DBLP:books/wi/HosmerL00}. Indeed, the inverse of logistic regression can be understood as a mapping from probability to a real-valued score. In logistic regression, when a probability $p_+ \in (0.5, 1)$, we consider it as positive. This condition is equivalent to inspecting whether the probability satisfies $p_+ > 1-p_+$, \ie, inspecting whether the odds satisfies $p_+/(1-p_+) \in (1, \infty)$. For this condition, the logarithm function maps the odds into a positive interval $\log{(p_+/(1-p_+))} \in (0, \infty)$. Similarly, for the negative, the condition $p_- \in (0, 0.5)$ is equivalent to $\log{(p_-/(1-p_-))} \in (-\infty, 0)$. In other words, the classification becomes inspecting the sign of the log-odds, which is modeled as a real-valued score such as an output of a fully connected layer. The log-odds function, which is also referred to as the logit function, maps a probability $p \in (0, 1)$ into the interval $(-\infty, \infty)$ with infinite length as $\logit(p) \coloneqq \log(p/(1-p))$, which is the inverse of the sigmoid function. This interpretation enables building logistic regression via the pipeline of a fully connected layer and a sigmoid function.

We leverage this log-odds function to map the cosine similarities into the input of softmax to replace the division by temperature scheme. Given $\cos{\theta_{ij}} \in (-1, 1)$, we cast it into a probability-like score via $p = (1+\cos{\theta_{ij}})/2 \in (0, 1)$. A cosine similarity for a positive pair is $\cos{\theta_{ij+} \in (0, 1)}$, which corresponds to $p_+ \in (0.5, 1)$ and thereby positive log-odds $\log(p_+/(1-p_+)) \in (0, \infty)$. Similarly, a cosine similarity for a negative pair is mapped into negative log-odds. The log-odds function for scaled cosine similarity is
\begin{align}
	\logit\left(\frac{1+\cos{\theta_{ij}}}{2}\right) = \log{\frac{1+\cos{\theta_{ij}}}{1-\cos{\theta_{ij}}}},
\end{align}
which ensures that its output is at interval $(-\infty, \infty)$ with infinite length. This infinite length of the interval addresses the representational limit in the division by temperature scheme. In other words, we map cosine similarities using the scaled log-odds function and feed them into the softmax in the InfoNCE loss, removing division by temperature.

Note that this function is equivalent to twice the inverse hyperbolic tangent function, \ie, $2 \arctanh{(\cos{\theta_{ij}})}$. A PyTorch built-in function of \texttt{torch.atanh} enables us to easily implement this function by injecting few lines of code into the existing design pattern of InfoNCE loss. An implementation example is presented in the Appendix.

In contrast to the division by temperature scheme, which causes vanishing gradients, we find that the proposed method is safe from this issue: The zero gradient occurs only at the optimal point, and other points exhibit alive gradients.

\begin{figure}[t!]
	\begin{center}
		\centerline{\includegraphics[width=\columnwidth]{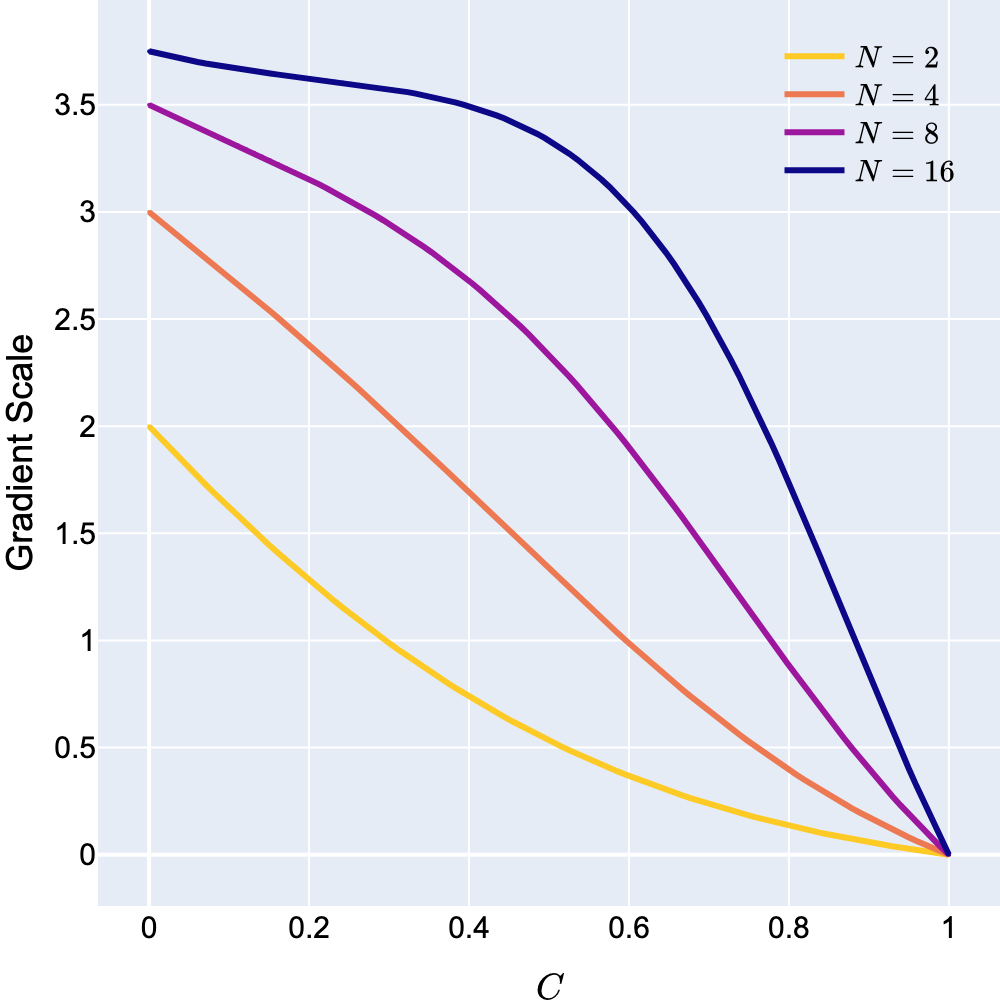}}
		\caption{Gradient scale corresponding to \eqref{eq:proposed}. The proposed method ensures the zero gradient when approaching the optimal point $C \rightarrow 1$ and nonzero gradient scales on other points with a monotonically decreasing function.}
		\label{fig:proposed}
	\end{center}
	\vskip -0.2in
\end{figure}

\begin{example}
	Imagine $N$ cosine similarities $[C, -C, \cdots, -C]$ with one positive and $N-1$ negatives, where $C \in (0, 1)$. In this scenario, the $i$th InfoNCE loss using our scaled log-odds mapping becomes
	\begin{align}
		L_i = -\log{\frac{(1+C)^2}{(1+C)^2 + (N-1)(1-C)^2}},
	\end{align}
	whose gradient with respect to $C$ is regulated by a scale factor of
	\begin{align}
		\abs{\pdv{L_i}{C}} = \frac{4(N-1)(1-C)}{(1+C)(N(1-C)^2 + 4C)}. \label{eq:proposed}
	\end{align}
	Note that in our scheme, as $C \rightarrow 1$, we have both $L_i \rightarrow 0$ and $\abs{\pdv{L_i}{C}} \rightarrow 0$ for any choice of $N$.

	\figref{fig:proposed} represents our gradient scale factor for $C \in (0, 1)$. The gradient scale becomes zero only when approaching the optimal point of $C \rightarrow 1$, whereas other points exhibit nonzero gradient scale factors as a monotonically decreasing function. This gradient behavior was consistently observed for $N \in \{2, 4, 8, 16\}$. On this basis, we claim that the proposed function provides a clear advantage in gradient descent compared with the division by temperature scheme.
\end{example}

\section{Experiments}
\label{sec:exp}
We now examine the performance of the proposed method across five benchmarks for contrastive learning. The performance was compared with that of the division by temperature scheme with several choices of temperatures.

\begin{table}[t]
	\caption{kNN top-1 accuracy (\%) for image classification on the Imagenette dataset. The underlined and bold values indicate the best results within division by temperature scheme and within all results, respectively.}
	\label{tab:ic}
	\begin{center}
		\begin{small}
			\begin{sc}
				\begin{tabular}{l|ccccc}
					\toprule
					$\tau$ & 0.1    & 0.25              & 0.5    & 1      & Free           \\
					\midrule
					Acc.   & 80.69  & \underline{84.43} & 83.75  & 82.27  & \textbf{84.65} \\
					Std.   & (0.24) & (0.20)            & (0.28) & (1.42) & (0.27)         \\
					\bottomrule
				\end{tabular}
			\end{sc}
		\end{small}
	\end{center}
	\vskip -0.1in
\end{table}

\subsection{Image Classification}
We first target image classification with contrastive learning, using the Imagenette dataset. The Imagenette dataset \citep{Howard_Imagenette_2019} is a subset of ten classes from ImageNet. A crop size of 128 pixels was applied after a resize operation using a size of 160 pixels. The objective was to train ResNet-18 \citep{DBLP:conf/cvpr/HeZRS16} from scratch using SimCLR and its NT-Xent loss \citep{DBLP:conf/icml/ChenK0H20}. To follow common practice for contrastive learning, training recipes from LightlySSL were employed. For training, the number of epochs of 800, stochastic gradient descent with a momentum of 0.9, learning rate of $6 \times 10^{-2}$ with cosine annealing scheduler \citep{DBLP:conf/iclr/LoshchilovH17}, weight decay of $5 \times 10^{-4}$, and mini-batch size of 256 were used. The training was performed on a single GPU machine. We measured the kNN top-1 accuracy (\%) and reported the average of ten runs with different random seeds (\tabref{tab:ic}).

Firstly, we observed that the division by temperature scheme was significantly sensitive to the choice of temperature; certain choices of $\tau=0.1$ or 1 decreased the top-1 accuracy. Furthermore, though the default temperature of SimCLR is $\tau=0.5$, we found that the other choice of $\tau=0.25$ rather improved performance. For this task, our method yielded an accuracy of 84.65\%, which is a slightly improved performance compared with the accuracy of 84.43\% from the best temperature.

\subsection{Graph Representation Learning}
Here, we target a graph representation learning task. Specifically, we assessed the quality of the learned node embeddings in the node classification task on the CiteSeer dataset. The CiteSeer dataset \citep{DBLP:conf/dl/GilesBL98} contains citation networks of scientific publications labeled in seven classes. We target the GRACE model \citep{DBLP:journals/corr/abs-2006-04131}, which adopts contrastive learning with the division by temperature scheme. Following the training recipe with the PyGCL library \citep{DBLP:conf/nips/0001XLW21}, for training, the number of epochs of 1000 and the Adam optimizer \citep{DBLP:journals/corr/KingmaB14} with learning rate of 0.01 were used. We measured micro- and marco-averaged F1-scores (\%) and reported the average of twenty runs with different random seeds (\tabref{tab:cs})

\begin{table}[t]
	\caption{Micro- and macro-averaged F1-scores (\%) for node classification using the GRACE model on the CiteSeer dataset. The values in parentheses represent the standard deviations across twenty runs. Higher is better.}
	\label{tab:cs}
	\begin{center}
		\begin{small}
			\begin{sc}
				\begin{tabular}{l|cc}
					\toprule
					Temperature & F1-Micro                 & F1-Macro                 \\
					\midrule
					0.1         & 58.94             (3.17) & 53.41             (2.72) \\
					0.25        & 62.50             (2.92) & 56.73             (2.87) \\
					0.5         & \underline{67.33} (3.02) & \underline{60.47} (2.82) \\
					1           & 67.05             (3.02) & 60.03             (2.47) \\
					Free (Ours) & \textbf{67.95}    (2.10) & \textbf{60.56}    (2.09) \\
					\bottomrule
				\end{tabular}
			\end{sc}
		\end{small}
	\end{center}
	\vskip -0.1in
\end{table}

\begin{table*}[t]
	\caption{ROC-AUC (\%) for anomaly detection using the MSC method on the CIFAR-10 dataset. The best temperature even differed by class within the division by temperature scheme.}
	\label{tab:ano}
	\begin{center}
		\begin{small}
			\begin{sc}
				\begin{tabular}{l|ccccc|c}
					\toprule
					Class & airplane                   & automobile                          & bird                                & cat                        & deer                       &                    \\
					\midrule
					0.1   & 96.898 (0.019)             & 98.537 (0.003)                      & 94.225 (0.026)                      & 93.683 (0.025)             & 96.666 (0.016)             &                    \\
					0.25  & \underline{97.018} (0.011) & \underline{98.656} (0.003)          & 94.804 (0.014)                      & \underline{94.291} (0.020) & 96.861 (0.010)             &                    \\
					0.5   & 96.888 (0.005)             & 98.633 (0.002)                      & \textbf{\underline{94.814}} (0.044) & 94.163 (0.013)             & \underline{96.863} (0.009) &                    \\
					1     & 96.712 (0.006)             & 98.548 (0.003)                      & 94.480 (0.021)                      & 93.739 (0.014)             & 96.700 (0.007)             &                    \\
					Free  & \textbf{97.145} (0.011)    & \textbf{98.713} (0.003)             & 94.406 (0.027)                      & \textbf{94.461} (0.031)    & \textbf{96.919} (0.006)    &                    \\
					\midrule
					Class & dog                        & frog                                & horse                               & ship                       & truck                      & mean               \\
					\midrule
					0.1   & 96.624 (0.014)             & 98.043 (0.017)                      & 98.215 (0.015)                      & 98.006 (0.008)             & 98.118 (0.013)             & 96.902             \\
					0.25  & \underline{97.168} (0.006) & \textbf{\underline{98.223}} (0.005) & \textbf{\underline{98.332}} (0.003) & 98.527 (0.004)             & \underline{98.274} (0.005) & \underline{97.215} \\
					0.5   & 97.166 (0.005)             & 98.223 (0.006)                      & 98.317 (0.005)                      & \underline{98.528} (0.006) & 98.242 (0.006)             & 97.184             \\
					1     & 96.997 (0.007)             & 98.156 (0.004)                      & 98.218 (0.003)                      & 98.406 (0.006)             & 98.160 (0.003)             & 97.012             \\
					Free  & \textbf{97.392} (0.006)    & 98.171 (0.013)                      & 98.281 (0.007)                      & \textbf{98.809} (0.010)    & \textbf{98.491} (0.005)    & \textbf{97.279}    \\
					\bottomrule
				\end{tabular}
			\end{sc}
		\end{small}
	\end{center}
	\vskip -0.1in
\end{table*}

Again, we observed that the division by temperature scheme was significantly sensitive to the choice of temperature. The best result within the division by temperature scheme was observed for $\tau=0.5$, which differs from that of the previous experiment. Here, replacing division by temperature with the proposed method yielded improved performance.

\subsection{Anomaly Detection}
Now, we visit anomaly detection using contrastive learning. We target the MSC method \citep{DBLP:conf/aaai/ReissH23}, which is a modified contrastive loss for anomaly detection. Using the CIFAR-10 dataset \citep{krizhevsky2009learning}, we regarded a single CIFAR-10 class as anomalous, whereas others were considered normal. For training, ResNet-152 \citep{DBLP:conf/cvpr/HeZRS16}, the number of epochs of 25, stochastic gradient descent with learning rate $10^{-5}$, weight decay of $5 \times 10^{-5}$, and mini-batch size of 32 were used. We measured the mean ROC-AUC (\%) and reported the average of five runs with different random seeds (\tabref{tab:ano}).

For the division by temperature scheme, the best temperature was 0.25 or 0.5; however, we observed that the best temperature even differed within this task among the ten classes in the CIFAR-10 dataset. Compared with these results, the proposed method outperformed for seven out of the ten classes. When measuring the mean ROC-AUC across the ten classes, our method presented improved performance.

\begin{table}[t]
	\caption{Results on MABEL. For the language modeling score (LM) and the idealized context association test (ICAT) score, higher is better. For the stereotype score (SS) with $\Diamond$, closer to 50 is better.}
	\label{tab:mab}
	\begin{center}
		\begin{small}
			\begin{sc}
				\begin{tabular}{ll|cc}
					\toprule
					Dataset                    & Metrics          & Baseline      & Ours                \\
					\midrule
					\multirow{3}{*}{StereoSet} & LM $\uparrow$    & 80.6    (0.5) & \textbf{81.0} (0.6) \\
					                           & SS $\Diamond$    & 55.8    (0.9) & \textbf{55.7} (1.1) \\
					                           & ICAT  $\uparrow$ & 71.3    (1.3) & \textbf{71.7} (1.7) \\
					\midrule
					CrowS-Pairs                & SS $\Diamond$    & 52.5    (1.9) & \textbf{52.3} (1.4) \\
					\bottomrule
				\end{tabular}
			\end{sc}
		\end{small}
	\end{center}
	\vskip -0.1in
\end{table}

\subsection{Further Results}
We provide additional results on natural language processing using MABEL \citep{DBLP:conf/emnlp/HeXFC22} in \tabref{tab:mab} and on sequential recommendation using DCRec \citep{DBLP:conf/www/YangHXHLL23} in \tabref{tab:dcr}. We used the StereoSet \citep{DBLP:conf/acl/NadeemBR20}, CrowS-Pairs \citep{DBLP:conf/emnlp/NangiaVBB20}, and MovieLens-20M \citep{DBLP:journals/tiis/HarperK16} datasets for these experiments. For training MABEL, the number of epochs of 2, learning rate of $5 \times 10^{-5}$, mini-batch size of 16, FP16 precision, maximum sequence length of 128, and temperature of 0.05 were used. For training DCRec, mini-batch size of 512, dropout probability of 0.5 for graph, hidden nodes, and attention, top co-interaction size of 4, mean of conformity weights of 0.4, coefficient for contrastive loss of 0.1, coefficient for KL divergence of 0.01, and temperature of 0.8 were used. We reported the average and standard deviation from ten runs with different random seeds. The baseline here indicates the performance of the division by temperature scheme using the reported value of temperature. Compared with the baseline, the proposed method yielded consistent improvements.

\begin{table}[t]
	\caption{Hit ratio (HR@N) and normalized discounted cumulative gain (NDCG@N) using DCRec. Higher is better.}
	\label{tab:dcr}
	\begin{center}
		\begin{small}
			\begin{sc}
				\begin{tabular}{l|cc}
					\toprule
					Metrics & Baseline          & Ours            \\
					\midrule
                    HR@1    & 0.1336   (0.0012) & \textbf{0.1360} (0.0004) \\
					HR@5    & 0.3719   (0.0010) & \textbf{0.3741} (0.0015) \\
					HR@10   & 0.5191   (0.0008) & \textbf{0.5209} (0.0018) \\
					ndcg@1  & 0.1336   (0.0012) & \textbf{0.1360} (0.0004) \\
					ndcg@5  & 0.2556   (0.0007) & \textbf{0.2579} (0.0007) \\
					ndcg@10 & 0.3031   (0.0008) & \textbf{0.3052} (0.0008) \\
					\bottomrule
				\end{tabular}
			\end{sc}
		\end{small}
	\end{center}
	\vskip -0.1in
\end{table}

\section{Conclusion}
This study challenges the division by temperature scheme used in InfoNCE loss for contrastive learning. We discovered that the division by temperature scheme exhibits a trade-off between alive gradients at nonoptimal points and convergence at the optimal point, whose compromise requires precise tuning of the temperature. We found that this behavior is also affected by other factors such as the number of pairs, which connotes that a universally applicable temperature for InfoNCE loss does not exist. To address this problem, we proposed a novel method that uses the inverse hyperbolic tangent function instead of division by temperature. The proposed method is hyperparameter-free, easy to deploy in existing source code of InfoNCE loss, and provides improved theoretical properties for gradient descent. Extensive experiments demonstrated that the performance of contrastive learning with the division by temperature scheme is significantly sensitive to temperature, whereas the proposed method yields comparable or even better performance without temperature tuning.

\section*{Impact Statement}
This paper presents work whose goal is to advance the field of Machine Learning. There are many potential societal consequences of our work, none which we feel must be specifically highlighted here.


\bibliography{example_paper}
\bibliographystyle{icml2025}

\end{document}